\documentclass[letterpaper, 10 pt, conference]{ieeeconf}

\IEEEoverridecommandlockouts                              
\usepackage{graphics} 
\usepackage[colorlinks=true,allcolors=blue]{hyperref}
\usepackage{stmaryrd} 
\usepackage{epsfig} 
\usepackage{times} 
\usepackage{amsmath} 
\usepackage{amssymb}  
\usepackage{float}
\usepackage{hyperref}
\usepackage{subcaption}
\usepackage{physics}
\usepackage{tikz}
\usepackage{pgfplots}
\usepackage{balance}
\usepackage{tabu}
\usepackage{xcolor}
\usepackage{multirow,booktabs}
\usepackage{cite}
\usepackage[linesnumbered,ruled]{algorithm2e}

\makeatletter
\newcommand*\bigcdot{\mathpalette\bigcdot@{.64}}
\newcommand*\bigcdot@[2]{\mathbin{\vcenter{\hbox{\scalebox{#2}{$\m@th#1\bullet$}}}}}
\makeatother

\newcommand{\sref}[1]{{Section \ref{#1}}}
\newcommand{\fref}[1]{Fig. \ref{#1}}
\newcommand{\tref}[1]{Table \ref{#1}}

\title{\LARGE \bf
F-LOAM : Fast LiDAR Odometry and Mapping}

\author{Han Wang, Chen Wang, Chun-Lin Chen, and Lihua Xie, \textit{Fellow}, IEEE
\thanks{
The work is supported by Delta-NTU Corporate Laboratory for Cyber-Physical Systems under the National Research Foundation Corporate Lab @ University Scheme. The demonstration video can be found at \url{https://youtu.be/QvXN5XhAYYw}.  
}
\thanks{Han Wang, Chun-Lin Chen, and Lihua Xie are with the School of Electrical and Electronic Engineering,
Nanyang Technological University, 50 Nanyang Avenue, Singapore 639798.
        {\tt\small e-mail: \{wang.han,ChenCL,elhxie\}@ntu.edu.sg}}
\thanks{Chen Wang is with the Robotics Institute, Carnegie Mellon University, Pittsburgh, PA 15213, USA. {\tt\small e-mail: chenwang@dr.com}}
}

\begin{document}
 
\maketitle
\thispagestyle{empty}
\pagestyle{empty}

\begin{abstract}

Simultaneous Localization and Mapping (SLAM) has wide robotic applications such as autonomous driving and unmanned aerial vehicles.
Both computational efficiency and localization accuracy are of great importance towards a good SLAM system. Existing works on LiDAR based SLAM often formulate the problem as two modules: scan-to-scan match and scan-to-map refinement. Both modules are solved by iterative calculation which are computationally expensive. 
In this paper, we propose a general solution that aims to provide a computationally efficient and accurate framework for LiDAR based SLAM. 
Specifically, we adopt a non-iterative two-stage distortion compensation method to reduce the computational cost. For each scan input, the edge and planar features are extracted and matched to a local edge map and a local plane map separately, where the local smoothness is also considered for iterative pose optimization. 
Thorough experiments are performed to evaluate its performance in challenging scenarios, including localization for a warehouse Automated Guided Vehicle (AGV) and a public dataset on autonomous driving. 
The proposed method achieves a competitive localization accuracy with a processing rate of more than 10 Hz in the public dataset evaluation, which provides a good trade-off between performance and computational cost for practical applications.
It is one of the most accurate and fastest open-sourced SLAM systems\footnote{\url{https://github.com/wh200720041/floam}} in KITTI dataset ranking.

\end{abstract}
\section{INTRODUCTION}
Simultaneous Localization and Mapping (SLAM) is one of the most fundamental research topics in robotics. It is the task to localize the robot and build the surrounding map in an unknown or partially unknown environment based on on-board sensors.
According to the perceptual devices, it can be roughly categorized as LiDAR and Visual SLAM. 
Compared to visual SLAM, LiDAR SLAM is often more accurate in pose estimation and is robust to environmental variations such as illumination and weather change \cite{debeunne2020review}. Therefore, LiDAR SLAM is widely adopted by many robotic applications such as autonomous driving \cite{milz2018visual}, drone inspection \cite{cunha2018ultra}, and warehouse manipulation \cite{ito2018small}.

\begin{figure}[t]
\begin{center}
\vspace{8pt}
\includegraphics[width=0.99\linewidth]{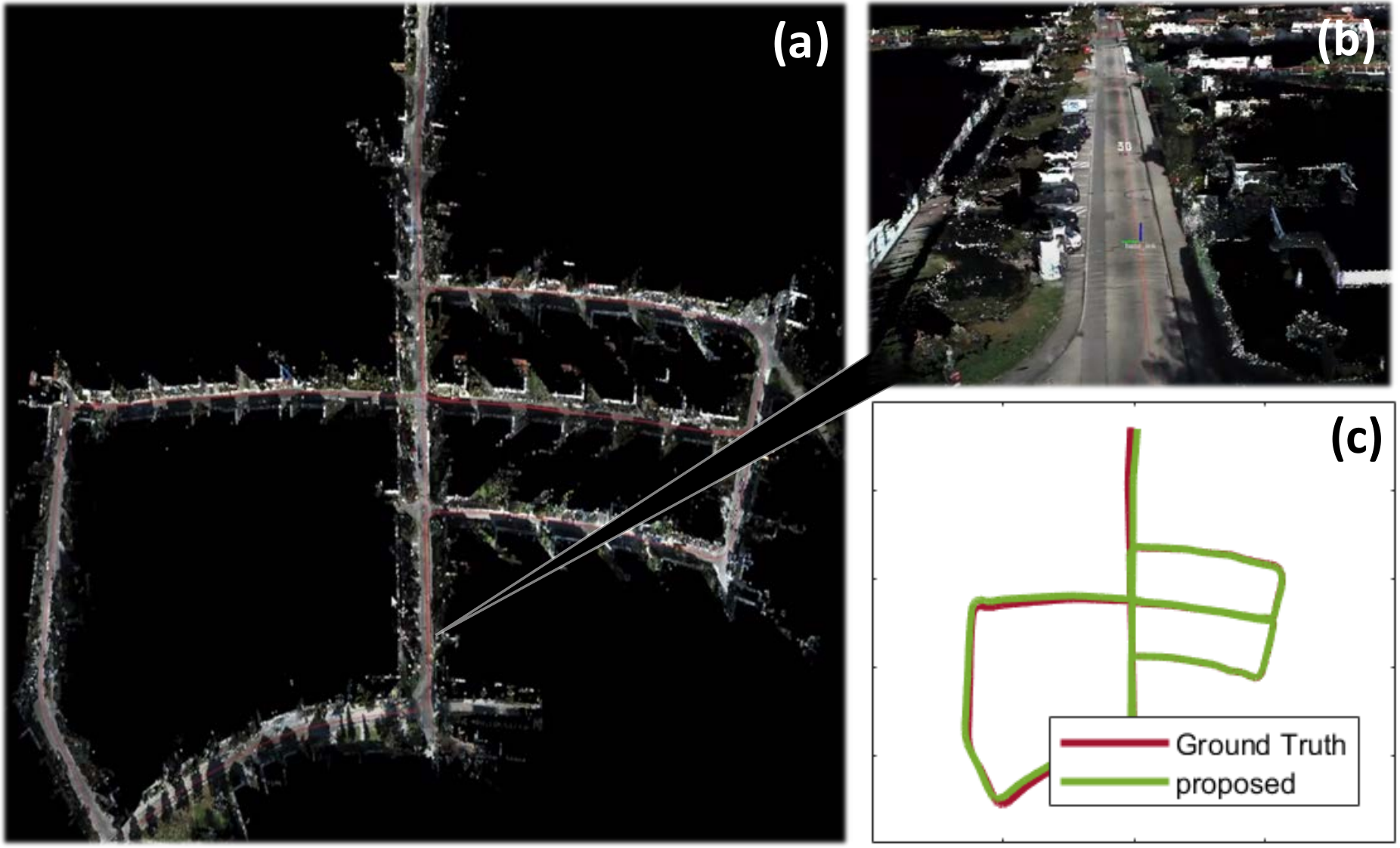}
\captionsetup{justification=justified}
\caption{Example of the proposed method on KITTI dataset. (a) shows the mapping result on sequence 05. (b) is the reconstructed 3D road scenery by integrating camera view. (c) plots the trajectory from F-LOAM and the ground truth.} 
\label{fig: title_graph}

\end{center}
\end{figure}


Although existing works on LiDAR SLAM have achieved good performance in public dataset evaluation, there are still some limitations in practical applications. One of the limitations is the robustness over different environments, \textit{e.g.}, from indoor to outdoor environments and from static to dynamic environments. For example, \cite{hess2016real} showed good performance in indoor scenario, but the localization accuracy drops a lot in outdoor environment. Another challenge is the computational cost. In many robotic platforms such as UAVs, the computational resources are limited, where the on-board processing unit is supposed to perform high frequency localization and path planning at the same time \cite{li2014lidar}. 
Moreover, many works are not open-sourced thus it is difficult to retrieve the same result. 

The most classical approach to estimate the transform between two scans is Iterative Closest Point (ICP) \cite{besl1992method}, where the two scans are aligned iteratively by minimizing point cloud distance. However, a large number of points are involved in optimization, which is computationally inefficient. Another approach is to match features that is more computationally efficient. A typical example is Lidar Odometry And Mapping (LOAM) \cite{zhang2017low} that extracts edge and planar features and calculates the pose by minimizing point-to-plane and point-to-edge distance. However, both distortion compensation and laser odometry require iterative calculation which are still computationally expensive. 

In this paper, we introduce a lightweight LiDAR SLAM that targets to provide a practical real-time LiDAR SLAM solution to public. A novel framework is presented that combines feature extraction, distortion compensation, pose optimization, and mapping. Compared to traditional method, we use a non-iterative two-stage distortion compensation method to replace the computationally inefficient iterative distortion compensation method. It is observed that edge features with higher local smoothness and planar features with lower smoothness are often consistently extracted over consecutive scans. Those points are more important for matching. Hence the local geometry feature is also considered for the iterative pose estimation to improve the localization accuracy. 
It is able to achieve real time performance up to 20 Hz on a low power embedded computing unit.
To demonstrate the robustness, a thorough evaluation of the proposed method is presented, including both indoor and outdoor experiments. Compared to existing state-of-the-art methods, our method is able to achieve competitive localization accuracy at a low computational cost, which is a good trade-off between performance and speed.
It is worthy noting that the proposed method is one of the most accurate and fastest open-sourced methods in the KITTI benchmark. 


This paper is organized as follows: \sref{sec:related-work} reviews the related work on existing LiDAR SLAM approaches. \sref{sec:methodology} describes the details of the proposed approach, including feature point selection, laser points alignment, laser odometry estimation, and mapping. \sref{sec:experiment} shows experiment results, followed by conclusion in \sref{sec:conclusion}.

\section{RELATED WORK}
\label{sec:related-work}
The most essential step in LiDAR SLAM is the matching of point clouds. Existing works mainly leverage on finding the correspondence between point cloud, including two main categories: raw point cloud matching and feature point pairs matching. For raw point cloud matching, the Iterative Closest Points (ICP) \cite{besl1992method} method is the most classic method. ICP measures the correspondent points by finding the closest point in Euclidean space. By iteratively minimizing the distance residual between corresponding points, the pose transition between two point clouds converges to the final location. An improved point cloud matching is IMLS-SLAM \cite{deschaud2018imls}, where a weight value is assigned to each point. Each weight is derived by IMplicit Least Square (IMLS) method based on the local surface normal of target point. However, it is often computationally costly to apply raw point cloud matching, \textit{e.g.}, it takes 1.2 s to estimate one frame in IMLS-SLAM that is far from real-time performance requirement. 

The feature-based methods mainly leverage on point-to-surface/edge matching and are popularly used. Introduced by LOAM \cite{zhang2014loam}, this has become the standard for the following work. The edges and surfaces are extracted in advance based on local smoothness analysis, and the points from the current point cloud are matched to the edges and surfaces from the map. The distance cost is formulated as the euclidean distance between the points and edges and surfaces. There are also many works that extends LOAM to achieve better performance, \textit{e.g.}, LeGO-LOAM separates ground optimization before feature extraction. For Autonomous Guided Vehicles (AGVs) \cite{shan2018lego}, the z-axis is not important and most of case we can assume robot moving in 2D space. Hence, ground points are not contributing to the localization of x-y plane. This approach is also widely implemented in ground vehicles. 


Some works attempt to improve the performance by introducing extra modules.
Loop closure detection is another key component in SLAM. The LiDAR odometry often comes with inevitable drifts which can be significant in long-term SLAM. Hence in some large scale scenarios, the loop closure serves in the back-end to identify repetitive places. For example, in LiDAR-only Odometry and Localization (LOL) the author extends LiDAR odometry into a full SLAM with loop closure \cite{rozenberszki2020lol}. By closing the loop, odometry drifting error can be corrected. 
Fusion with Inertial Measurement Unit (IMU) is an alternative solution. 
Shan \textit{et al} propose a fusion method with IMU and GPS named LIO-SLAM \cite{liosam2020shan}. Different sensor inputs are synchronized and tightly-coupled. The fused SLAM system is proven to be more accurate than LOAM in outdoor environment.
Fusion-based approaches have shown improvement on the localization accuracy. However, multi-sensor fusion requires synchronization and comprehensive calibration.

In the past few years there are also some deep learning related works that attempt to use Convolution Neural Network (CNN) for point cloud processing. Rather than hand-crafted features, features are extracted by CNN training. For example, in CAE-LO \cite{yin2020cae}, the author proposes a deep-learning-based feature point selection rules via unsupervised Convolutional Auto-Encoder (CAE) \cite{masci2011stacked}, while the LiDAR odometry estimation is still based on feature points matching. The experiment result shows that the learnt features can improve the matching success rate. 
Although deep learning approaches have shown good performance in public dataset, it is not robust in different environment such as indoor/outdoor and city/rural scenarios. For example, \cite{shi2020spsequencenet} has a good performance in semantic KITTI dataset. However, the same algorithm fails when moving to indoor environment or in some complex environment. 

\section{METHODOLOGY}
\label{sec:methodology}
In this section, the proposed method is described in details. We firstly present the sensor model and feature extraction. The extracted features are then calibrated with distortion compensation. Then, we present general feature extraction and global feature estimation. Lastly we explain the calculation of laser odometry as well as mapping.

\subsection{Sensor Model and Feature Extraction}
A mechanical 3D LiDAR perceives the surrounding environment by rotating a vertically aligned laser beam array of size $M$. 
It scans vertical planes with $M$ readings in parallel. 
The laser array is rotating in a horizontal plane at constant speed during each scan interval, while the laser measurements are taken in clockwise or counterclockwise order sequentially \cite{bergelt2017improving}. 
For a mechanical LiDAR sensor model, we denote the $k^{th}$ LiDAR scan as $P_{k}$, and each point as $\textbf{p}_{k}^{(m,n)}$, where $m \in [1,M]$ and $n \in [1,N]$. 
Raw point cloud matching methods such as ICP are sensitive to noise and dynamic objects such as human for autonomous driving. Moreover, a LiDAR scan contains tens of thousands of points which makes ICP computationally inefficient. Compared to raw point cloud matching method such as ICP, feature point matching is more robust and efficient in practice. To improve the matching accuracy and matching efficiency, we leverage surface features and edge features, while discard those noisy or less significant points.  
As mentioned above, the point cloud returned by a 3D mechanical LiDAR is sparse in vertical direction and dense in horizontal direction. Therefore, horizontal features are more distinctive and it is less likely to have false feature detection in horizontal plane. For each point cloud, we focus on horizontal plane and evaluate the smoothness of local surface by 
\begin{equation}
  \sigma_{k}^{(m,n)} = \frac{1}{ \left| \mathcal{S}_{k}^{(m,n)} \right| } \sum_{\textbf{p}_{k}^{(m,j)} \in \mathcal{S}_{k}^{(m,n)}}(||\textbf{p}_{k}^{(m,j)} - \textbf{p}_{k}^{(m,n)}||),
\end{equation}
where $\mathcal{S}_{k}^{(m,n)}$ is the adjacent points of $\textbf{p}_{k}^{(m,n)}$ in horizontal direction and $|\mathcal{S}_{k}^{(m,n)}|$ is the number of points in local point cloud. $\mathcal{S}_{k}^{(m,n)}$ can be easily collected based on point ID $n$, which can reduce the computational cost compared to local searching. In experiment we pick 5 points along the clockwise and counterclockwise, respectively. For a flat surface such as walls, the smoothness value is small while for corner or edge point, the smoothness value is large. For each scan plane $\mathcal{P}_{k}$, edge feature points are selected with high $\sigma$, and surface feature points are selected with low $\sigma$. Therefore, we can formulate the edge feature set as $\mathcal{E}_{k}$ and the surface feature set as $\mathcal{S}_{k}$. 

\subsection{Motion Estimation and Distortion Compensation} \label{sec: distortion}

In existing works such as LOAM \cite{zhang2017low} and LeGO-LOAM \cite{shan2018lego}, distortion is corrected by scan-to-scan match which iteratively estimates the transformation between two consecutive laser scans. 
However, iterative calculation is required to find the transformation matrix which is computationally inefficient. In this paper, we propose to use two-stage distortion compensation to reduce the computational cost. Note that most existing 3D LiDARs are able to run at more than 10 Hz and the time elapsed between two consecutive LiDAR scans is often very short. Hence we can firstly assume constant angular velocity and linear velocity during short period to predict the motion and correct the distortion. In the second stage, the distortion will be re-computed after the pose estimation process and the re-computed undistorted features will be updated to the final map. In the experiment we find that the two-stage distortion compensation can achieve the similar localization accuracy but at much less computational cost. Denote the robot's pose at $k^{th}$ scan as a 4x4 homogeneous transformation matrix $\textbf{T}_{k}$, the 6-DoF transform between two consecutive frames $k-1$ and $k$ can be estimated by:
\begin{equation}
  \xi^{k}_{k-1} = \log ({\textbf{T}_{k-2}}^{-1} \textbf{T}_{k-1}),
\end{equation}
where $\xi \in \mathfrak{se}(3)$.
The transform of small period $\delta t$ between the consecutive scans can be estimated by linear interpolation:
\begin{equation}
\label{eq: distortion compensation}
  \textbf{T}_{k}(\delta t) = \textbf{T}_{k-1} \exp(\frac{N-n}{N}\xi^{k}_{k-1}),
\end{equation}
where function $\exp(\xi)$ transforms a Lie algebra into Lie group defined by \cite{kirillov2008introduction}.
The distortion of current scan $\mathcal{P}_{k}$ can be corrected by:
\begin{equation}
  \tilde{\mathcal{P}}_{k} = \{ \textbf{T}_{k}(\delta t)\, \textbf{p}_{k}^{(m,n)}\,\,|\,\,\, \textbf{p}_{k}^{(m,n)} \in \mathcal{P}_{k}\}.
\end{equation}
The undistorted features will be used to find the robot pose in the next section. 
\subsection{Pose Estimation}
The pose estimation aligns the current undistorted edge features $\tilde{\mathcal{E}_k}$ and planar features $\tilde{\mathcal{S}_k}$ with the global feature map. The global feature map consists of a edge feature map and a planar feature map and they are updated and maintained separately. 
To reduce the searching computational cost, both edge and planar map are stored in 3D KD-trees. Similar to \cite{zhang2017robust}, the global line and plane are estimated by collecting nearby points from the edge and planar feature map. 
For each edge feature point $\textbf{p}_\mathcal{E} \in \tilde{\mathcal{E}_{k}}$, we compute the covariance matrix of its nearby points from the global edge feature map. When the points are distributed in a line, the covariance matrix contains one eigenvalue that is much larger. The eigenvector $\textbf{n}_\mathcal{E}^g$ associated with the largest eigenvalue is considered as the line orientation and the position of the line $\textbf{p}_\mathcal{E}^g$ is taken as the geometric center of the nearby points. Similarly, for each planar feature point $\textbf{p}_\mathcal{S} \in \tilde{\mathcal{S}_{k}}$, we can get a global plane with position $\textbf{p}_\mathcal{S}^g$ and surface norm $\textbf{n}_\mathcal{S}^g$. Note that different from global edge, the norm of global plane is taken as the eigenvector associated with the smallest eigenvalue.

\begin{figure*}[t]
\begin{center}
\vspace{4px}
\includegraphics[width=0.99\linewidth]{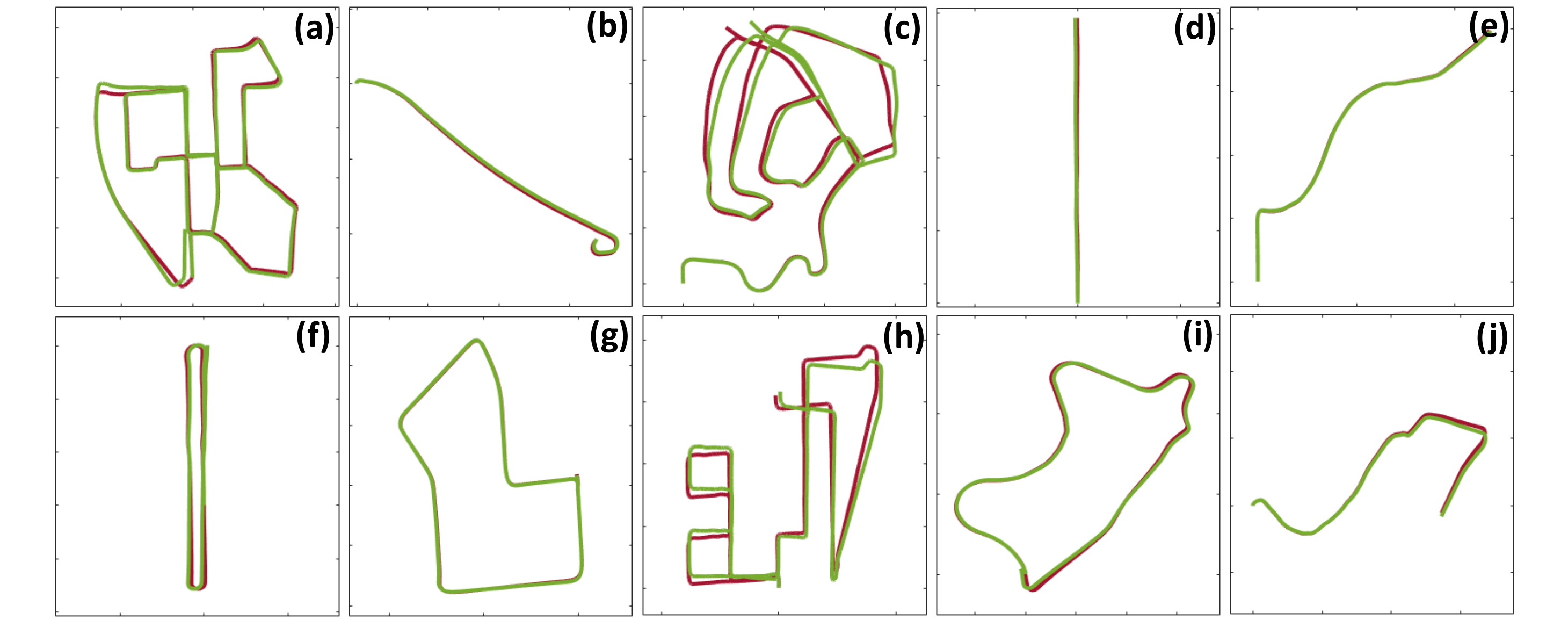}
\captionsetup{justification=justified}
\caption{Result of proposed F-LOAM on KITTI dataset. The estimated trajectory and ground truth are plotted in green and red color respectively. (a)-(e) sequence 00-04. (f)-(j): sequence 06-10.}
\label{fig:kittiresult}
\vspace{-10pt}
\end{center}
\end{figure*}

Once the corresponding optimal edges/planes are derived, we can find the linked global edges or planes for each feature points from $\tilde{\mathcal{P}}_{k}$. Such correspondence can be used to estimate the optimal pose between current frame and global map by minimizing the distance between feature points and global edges/planes. The distance between the edges features and global edge is:
\begin{equation}
f_{\mathcal{E}}(\textbf{p}_{\mathcal{E}}) = \textbf{p}_{\textbf{n}} \bigcdot \left((\textbf{T}_k \textbf{p}_{\mathcal{E}} - \textbf{p}_\mathcal{E}^g) \cross \textbf{n}_\mathcal{E}^g\right),
\end{equation}
where symbol $\bigcdot$ is the dot product and $\textbf{p}_{\textbf{n}}$ is the unit vector given by 
\begin{equation}
\textbf{p}_{\textbf{n}} = \frac{(\textbf{T}_k \textbf{p}_{\mathcal{E}} - \textbf{p}_\mathcal{E}^g) \cross \textbf{n}_\mathcal{E}^g}{||(\textbf{T}_k \textbf{p}_{\mathcal{E}} - \textbf{p}_\mathcal{E}^g) \cross \textbf{n}_\mathcal{E}^g||}.
\end{equation}
The distance between the planar features and global plane:
\begin{equation}
f_{\mathcal{S}}(\textbf{p}_{\mathcal{S}}) = (\textbf{T}_k  \textbf{p}_{\mathcal{S}} - \textbf{p}_\mathcal{S}^g) \bigcdot \textbf{n}_\mathcal{S}^g.
\end{equation}

Traditional feature matching only optimizes the geometry distance mentioned above, while the local geometry distribution of each feature point is not considered. However,  It is observed that edge features with higher local smoothness and planar features with lower smoothness are often consistently extracted over consecutive scans, which is more important for matching. Therefore, a weight function is introduced to further balance the matching process. To reduce the computational cost, the local smoothness defined previously is re-used to determine the weight function. For each edge point $\textbf{p}_{\mathcal{E}}$ with local smoothness $\sigma_{\mathcal{E}}$ and each plane point $\textbf{p}_{\mathcal{S}}$ with local smoothness $\sigma_{\mathcal{S}}$, the weight is defined by: 
\begin{equation}
\begin{aligned}
W(\textbf{p}_{\mathcal{E}}) &= \frac{\exp(-\sigma_{\mathcal{E}})}{\sum_{\textbf{p}^{(i,j)}\in \tilde{\mathcal{E}_{k}}} \exp(-\sigma^{(i,j)}_{k})}\\
W(\textbf{p}_{\mathcal{S}}) &= \frac{\exp(\sigma_{\mathcal{S}})}{\sum_{\textbf{p}^{(i,j)}\in \tilde{\mathcal{S}_{k}}} \exp(\sigma^{(i,j)}_{k})},
\end{aligned}
\end{equation}
where $\mathcal{E}_k$ and $\mathcal{S}_k$ are the edge features and plane features at $k^{th}$ scan.
The new pose is estimated by minimizing weighted sum of the point-to-edge and point-to-planar distance: 
\begin{equation}
\min_{\textbf{T}_{k}} \sum W(\textbf{p}_{\mathcal{E}}) f_{\mathcal{E}}(\textbf{p}_{\mathcal{E}}) + \sum W(\textbf{p}_{\mathcal{S}}) f_{\mathcal{S}}(\textbf{p}_{\mathcal{S}}).
\end{equation}
The optimal pose estimation can be derived by solving the non-linear equation through Gauss-Newton method. The Jacobian can be estimated by applying left perturbation model with $\delta\xi \in \mathfrak{se}(3)$ \cite{barfoot2016state}:
\begin{equation}
\begin{aligned}
 \textbf{J}_p = \frac{\partial \textbf{T}\textbf{p}}{\partial \delta\xi} & =\lim_{\delta \xi \to \textbf{0}} \frac{(\exp({\delta \xi)}\, \textbf{T}\textbf{p} - \textbf{T}\textbf{p})}{\delta \xi} \\
    &= \begin{bmatrix}
\mathbf{I}_{3\times 3}  &-[\textbf{T}\textbf{p}]_{\times}\\
\textbf{0}_{1\times 3} & \textbf{0}_{1\times 3}
\end{bmatrix},
\end{aligned}
\end{equation}
where $[\textbf{T}_k\textbf{p}_k]_{\times}$ transforms 4D point expression $\{x,y,z,1\}$ into 3D point expression $\{x,y,z\}$ and calculates its skew symmetric matrix. The Jacobian matrix of edge residual can be calculated by:
\begin{equation}
   \textbf{J}_{\mathcal{E}} = W(\textbf{p}_{\mathcal{E}})\, \frac{\partial f_{\mathcal{E}}}{\partial \textbf{T}\textbf{p}} \,\frac{\partial \textbf{T}\textbf{p}}{\partial \delta\xi} = W(\textbf{p}_{\mathcal{E}}) \, \textbf{p}_{\textbf{n}} \bigcdot (\textbf{n}_\mathcal{E}^g \times \textbf{J}_p),
\end{equation}
Similarly, we can derive
\begin{equation}
   \textbf{J}_{\mathcal{S}} = W(\textbf{p}_{\mathcal{S}})\, \frac{\partial f_{\mathcal{S}}}{\partial \textbf{T}\textbf{p}} \,\frac{\partial \textbf{T}\textbf{p}}{\partial \delta\xi} = W(\textbf{p}_{\mathcal{S}}) \, \textbf{n}_\mathcal{E}^g \bigcdot \textbf{J}_p .
\end{equation}
By solving nonlinear optimization we can derive odometry estimation based on above correspondence. Then, we can use this result to calculate new correspondence and new odometry. The current pose estimation $\mathbf{T}_k^*$ can be solved by iterative pose optimization until it converges.

\begin{figure*}[t]
\vspace{4px}
\begin{center}
\includegraphics[width=0.99\linewidth]{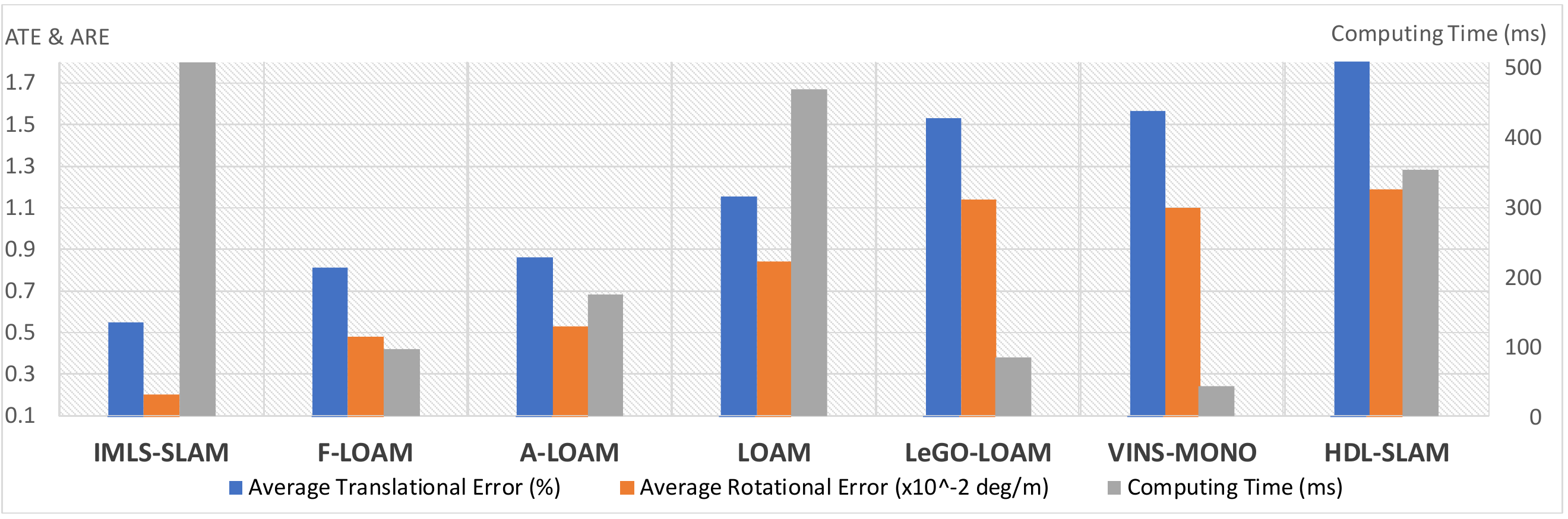}
\captionsetup{justification=justified}
\caption{Comparison of the different localization approaches on KITTI dataset sequence 00-10.}
\label{fig: kitticomparison}
\vspace{-10pt}
\end{center}
\end{figure*}

\subsection{Mapping Building \& Distortion Compensation Update}
The global map consists of a global edge map and a global planar map and is updated based on keyframes. Each keyframe is selected when the translational change is greater than a predefined translation threshold, or rotational change is greater than a predefined rotation threshold. The keyframe-based map update can reduce the computational cost compared to frame-by-frame update. As mentioned in \sref{sec: distortion}, the distortion compensation is performed based on the constant velocity model instead of iterative motion estimation in order to reduce the computational cost. However, this is less accurate than iterative distortion compensation in LOAM. Hence, in the second stage, the distortion is re-computed based on the optimization result $\textbf{T}_k^*$ from the last step by:
\begin{subequations}
\begin{align}
    \Delta \xi^* = \log &({\textbf{T}_{k-1}}^{-1} \cdot \textbf{T}_{k}^*), \\
    \tilde{\mathcal{P}}_{k}^* = \{ \exp(\frac{N-n}{N}\cdot \Delta \xi^*)\,& \textbf{p}_{k}^{(m,n)}\,\,|\,\,\, \textbf{p}_{k}^{(m,n)} \in \mathcal{P}_{k}\}.
\end{align}
\end{subequations}
The re-computed undistorted edge features and planar features will be updated to the global edge map and the global planar map respectively. After each update, the map is down-sampled by using a 3D voxelized grid approach \cite{rusu20113d} in order to prevent memory overflow.

\section{EXPERIMENT EVALUATION}
\label{sec:experiment}

\subsection{Experiment Setup}
To validate the algorithm, we evaluate F-LOAM on both a large scale outdoor environment and a medium scale indoor environment. For the large scale experiment, we evaluate our approach on KITTI dataset \cite{geiger2013vision} which is one of the most popular datasets for SLAM evaluation. Then the algorithm is integrated into warehouse logistics. It is firstly validated in a simulated warehouse environment and then is tested on an AGV platform.

\subsection{Evaluation on Public Dataset}
We firstly test our method on KITTI dataset \cite{Geiger2012CVPR} that is popularly used for outdoor localization evaluation. The dataset is collected from a driving car equipped with Velodyne HDL-64 LiDAR, cameras, and GPS. Most state-of-the-art SLAM methods have been evaluated on this dataset, \textit{e.g.}, ORB-SLAM \cite{mur2015orb}, VINS-Fusion \cite{qin2019general}, LIMO \cite{graeter2018limo}, and LSD-SLAM \cite{engel2015large}. To validate the robustness, we evaluate the proposed algorithm on all sequences from the KITTI dataset, which includes different scenarios such as high way, city center, county road, residential area, \textit{etc}. The ground truth of sequence 11-21 are not open to public and therefore, we mainly illustrate our performance on sequence 00-11.
We calculate the Average Translational Error (ATE) and Average Rotational Error (ARE) which are defined by the KITTI dataset \cite{Geiger2012CVPR}: 
\begin{equation}
\begin{aligned}
  E_{rot}(\mathcal{F}) &= \frac{1}{|\mathcal{F}|} \sum_{i,j \in \mathcal{F}} \angle[\hat{T}_j \hat{T}_i^{-1} T_i T_j^{-1} ]\\
  E_{trans}(\mathcal{F}) &= \frac{1}{|\mathcal{F}|} \sum_{i,j \in \mathcal{F}} {||\hat{T}_j \hat{T}_i^{-1} T_i T_j^{-1} ||}_2,\\
\end{aligned}
\end{equation}
where $\mathcal{F}$ is a set of frames $(i, j)$, $T$ and $\hat{T}$ are the estimated and true LiDAR poses respectively, $\angle[\cdot]$ is the rotation angle. The results are shown in \fref{fig:kittiresult}. Our methods achieves an average translational error of 0.80\% and an average rotational error of 0.0048 $\deg/m$ over 11 sequences.

\begin{figure}[t]
\begin{center}
\includegraphics[width=0.99\linewidth]{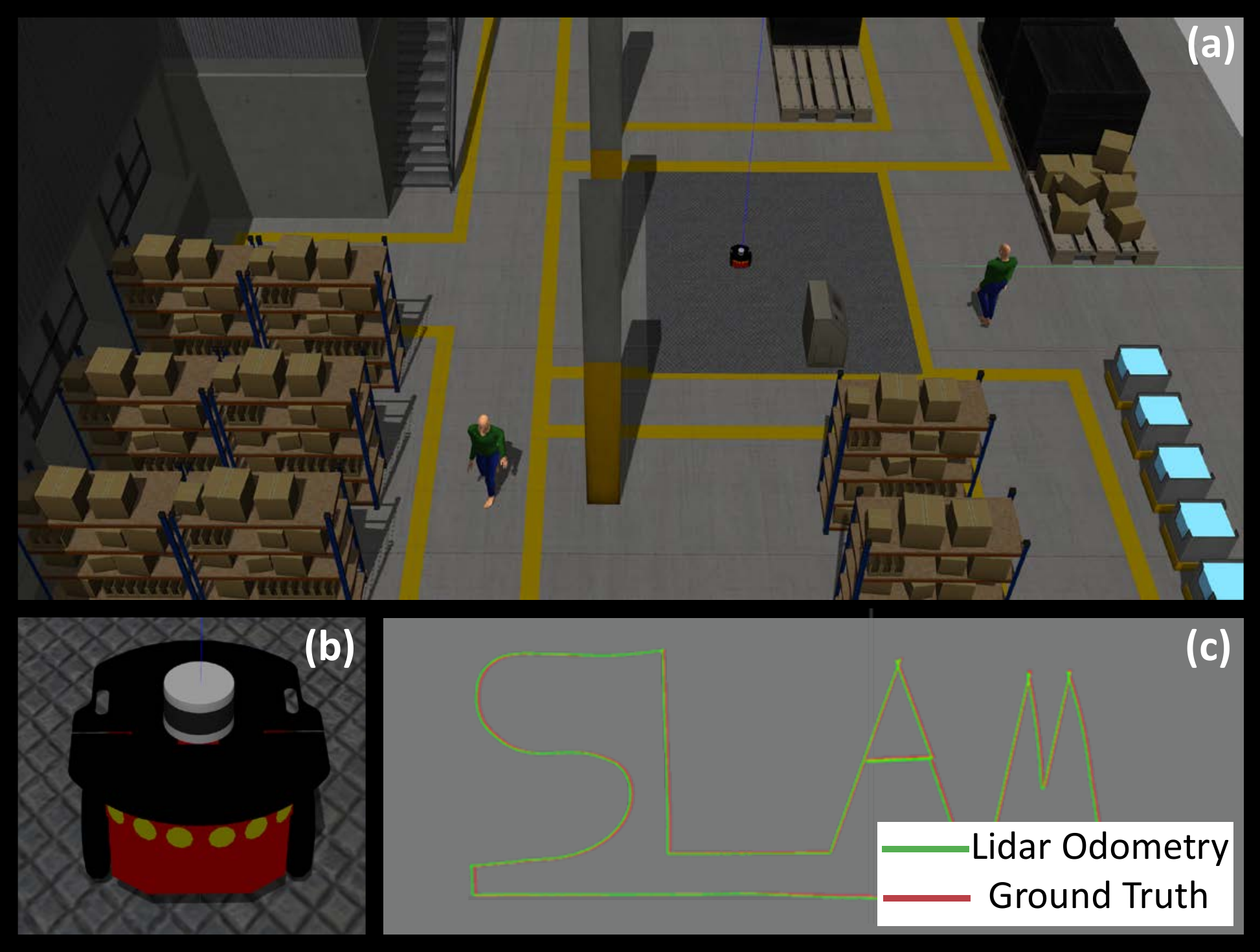}
\captionsetup{justification=justified}
\caption{Simulated warehouse environment with both static and dynamic objects. (a) Simulated environment in Gazebo. (b) Simulated Pioneer robot and Velodyne LiDAR for evaluation. (c) Trajectory comparison of F-LOAM and ground truth.}
\label{fig:simulation}
\vspace{-10pt}
\end{center}
\end{figure}

\begin{figure*}[t]
\vspace{4px}
\begin{center}
\includegraphics[width=0.99\linewidth]{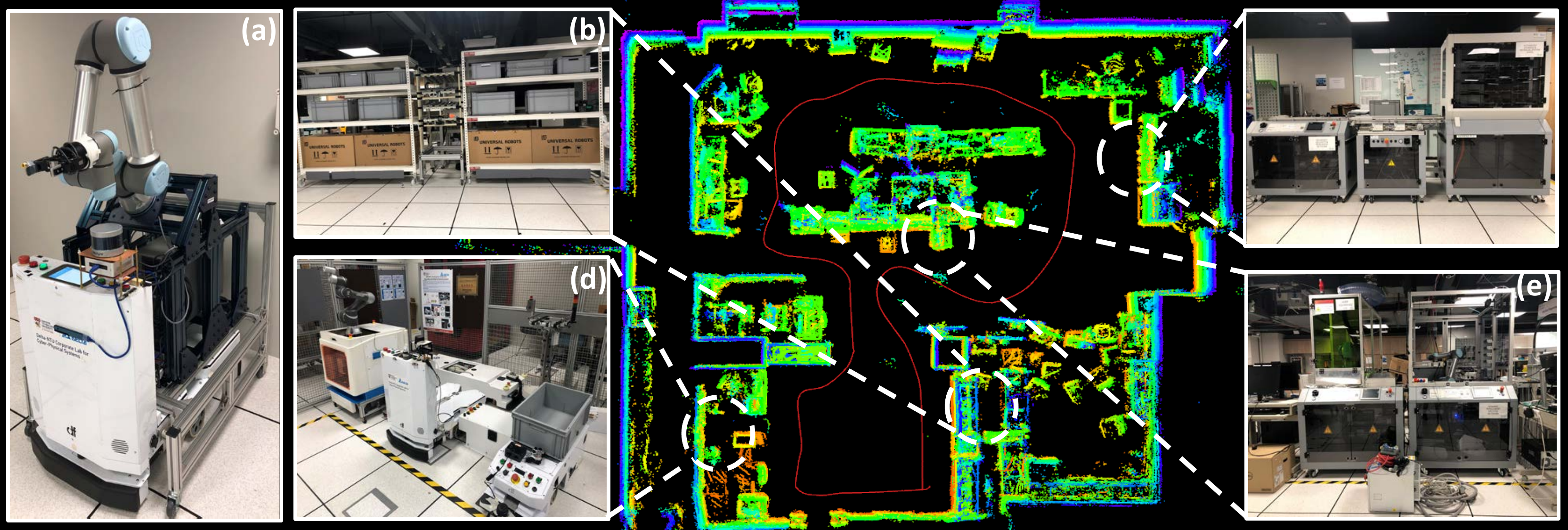}
\captionsetup{justification=justified}
\caption{F-LOAM in warehouse environment. (a) Automated Guided Vehicle used for experiment. (b-e) Advanced factory environment built for AGV manipulation, including operating machine, auto charging station, and storage shelves. Center image: F-LOAM result of warehouse localization and mapping. }
\label{fig:warehouse_robot}
\vspace{-10pt}
\end{center}
\end{figure*}

\begin{figure}[t]
\begin{center}
\includegraphics[width=0.89\linewidth]{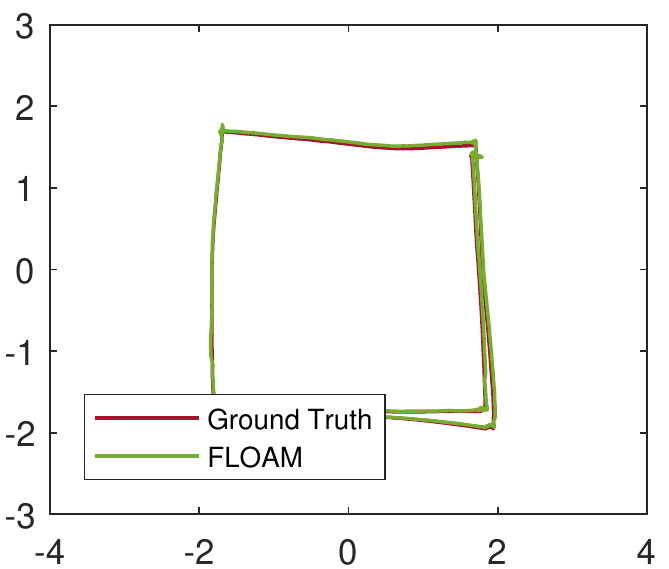}
\captionsetup{justification=justified}
\caption{Comparison of the proposed method and ground truth. Our method can precisely track robot's pose and it achieves an average localization error of 2 \textit{cm}}
\label{fig:viconcompare}
\vspace{-10pt}
\end{center}
\end{figure}

We also compare our method with state-of-the-art methods, including both LiDAR SLAM and visual SLAM.
In total there are 23,201 frames and traveled length of 22 \textit{km} over 11 sequences. The average computing time over all sequences is also recorded. In order to have precise evaluation on the computing time, all of the mentioned methods are tested on an Intel i7 3.2GHz processor based computer. The time is calculated from the start of feature extraction till the output of estimated odometry.
The proposed method is compared with the open-sourced state-of-the-art LiDAR methods such as LOAM \cite{zhang2014loam}, A-LOAM, HDL-Graph-SLAM \cite{koide2019portable}, IMLS-SLAM \cite{deschaud2018imls}, and LeGO-LOAM \cite{shan2018lego}. Noted that IMLS-SLAM is not open sourced and the results are collected from the original paper \cite{deschaud2018imls}. In additional, the visual SLAM such as VINS-MONO \cite{qin2018vins} is also compared. For consistency, the IMU information is not used and the loop closure detection is removed. The testing environment is based on ROS Melodic \cite{o2014gentle} and Ubuntu 18.04. The result is shown in \fref{fig: kitticomparison}. IMLS-SLAM achieves the highest accuracy. However, all points are used in iterative pose estimation and the processing speed is slow. In terms of computational cost, LeGO-LOAM is the fastest LiDAR SLAM since it only applies optimization on non-ground points. Among all methods compared, our method achieves second highest accuracy at an average processing rate of more than 10 Hz, which is a good trade-off between computational cost and localization accuracy. Our method is able to provide both fast and accurate SLAM solution for real time robotic applications and is competitive among state-of-the-art approaches.  


\subsection{Experiment on warehouse logistics}
In this experiment, we target to build an autonomous warehouse robot to replace human-labour-dominated manufacturing. The AGV is designed to carry out daily tasks such as transportation. This requires the robot platform to actively localize itself in a complex environment. 

\subsubsection{Simulation} We firstly validate our algorithm in a simulated environment. The simulation environment is built up on Gazebo and Linux Ubuntu 18.04. As shown in \fref{fig:simulation} (a), we use a virtual Pioneer robot and a virtual Velodyne VLP-16 as the ground vehicle platform. The simulated environment reconstructs a complex warehouse environment including various objects such as moving human workers, shelves, machines, \textit{etc}. 
The robot is controlled by a joystick to move around at the maximum speed of 2 $m/s$. We record the trajectory from F-LOAM and compare it with the ground truth. The result is shown in \fref{fig:simulation} (c). The F-LOAM trajectory and ground truth are plotted in green and red color respectively. Our method is able to track the AGV moving in high speed under the dynamic environment where the human operators are walking in the warehouse environment.

\subsubsection{Experiment} To further demonstrate the performance of our method, we implement F-LOAM on an actual AGV used for smart manufacturing. As shown in \fref{fig:warehouse_robot} (b-e), the warehouse environment consists of three main areas: auto charging station, material handling area and manufacturing station. A fully autonomous factory requires the robot to deliver material to manufacturing machines for assembly and collect products. All these operations require precise localization to ensure the reliability and safety. The robot platform for testing is shown in \fref{fig:warehouse_robot} (a) and is equipped with an Intel NUC mini computer and a Velodyne VLP-16 sensor. The localization and mapping result can be found in \fref{fig:warehouse_robot}. In this scenario, the robot automatically explores the warehouse and builds the map simultaneously using the proposed method.

\subsubsection{Performance evaluation} The proposed method is also tested in an indoor room equipped with a VICON system to evaluate its localization accuracy. The robot is remotely controlled to move in the testing area. The results are shown in \fref{fig:viconcompare}, where the F-LOAM trajectory and the ground truth trajectory are plotted in green and red color respectively. It can be seen that our method can accurately track the robot's pose. It achieves an average localization accuracy of 2 \textit{cm} compared to the ground truth provided by the VICON system.

\subsubsection{Ablation study} To further evaluate the performance of the proposed distortion compensation approach, we compare the results of different approaches in the warehouse environment. We firstly remove the distortion compensation in F-LOAM and record both computing time and localization accuracy. Then we add the iterative distortion compensation method from LOAM into our method and record the result. Lastly we use the proposed motion compensation and record the result. The results are shown in \tref{table:ablation study}. It can be seen that the proposed approach is much faster than the LOAM with motion compensation yet with a slightly better localization accuracy..
\begin{table}[t]
\vspace{6px}
    \begin{center}
    \begin{tabular}{ccc}
    \toprule
    \multirow{2}{*}{\textbf{Methods}}  & \textbf{Computing Time } & \textbf{Accuracy}  \\
    &\textbf{(ms/frame)}& \textbf{(cm)}\\
    \hline
    \midrule
    No Compensation & \textbf{66.33}    &  2.132  \\  
    LOAM & 84.52    &   2.052 \\  
    F-LOAM  & 69.07 & \textbf{2.037}    \\  
    
    \bottomrule
    \end{tabular}
    \caption{Ablation study of localization accuracy and computational cost.}
    \label{table:ablation study}
    \end{center}
\end{table}



\section{CONCLUSION}
\label{sec:conclusion}
In this paper, we present a computationally efficient LiDAR SLAM framework which targets to provide a public solution to robotic applications with limited computational resources. Compared to traditional methods, we propose to use a non-iterative two stage distortion compensation to reduce the computational cost. It is also observed that edge features with higher local smoothness and planar features with lower smoothness are often consistently extracted over consecutive scans, which are more important for scan-to-map matching. Therefore, the local geometry feature is also considered for iterative pose estimation. To demonstrate the robustness of the proposed method in practical applications, thorough experiments have been done to evaluate the performance, including simulation, indoor AGV test, and outdoor autonomous driving test. Our method achieves an average localization accuracy of 2 \textit{cm} in indoor test and is one of the most accurate and fastest open-sourced methods in KITTI dataset. 

\balance
\bibliographystyle{IEEEtran}
\bibliography{IEEEabrv,references}

\begin{thebibliography}{10}
\providecommand{\url}[1]{#1}
\csname url@rmstyle\endcsname
\providecommand{\newblock}{\relax}
\providecommand{\bibinfo}[2]{#2}
\providecommand\BIBentrySTDinterwordspacing{\spaceskip=0pt\relax}
\providecommand\BIBentryALTinterwordstretchfactor{4}
\providecommand\BIBentryALTinterwordspacing{\spaceskip=\fontdimen2\font plus
\BIBentryALTinterwordstretchfactor\fontdimen3\font minus
  \fontdimen4\font\relax}
\providecommand\BIBforeignlanguage[2]{{%
\expandafter\ifx\csname l@#1\endcsname\relax
\typeout{** WARNING: IEEEtran.bst: No hyphenation pattern has been}%
\typeout{** loaded for the language `#1'. Using the pattern for}%
\typeout{** the default language instead.}%
\else
\language=\csname l@#1\endcsname
\fi
#2}}

\bibitem{debeunne2020review}
C.~Debeunne and D.~Vivet, ``A review of visual-lidar fusion based simultaneous
  localization and mapping,'' \emph{Sensors}, vol.~20, no.~7, p. 2068, 2020.

\bibitem{milz2018visual}
S.~Milz, G.~Arbeiter, C.~Witt, B.~Abdallah, and S.~Yogamani, ``Visual slam for
  automated driving: Exploring the applications of deep learning,'' in
  \emph{Proceedings of the IEEE Conference on Computer Vision and Pattern
  Recognition Workshops}, 2018, pp. 247--257.

\bibitem{cunha2018ultra}
F.~Cunha and K.~Youcef-Toumi, ``Ultra-wideband radar for robust inspection
  drone in underground coal mines,'' in \emph{2018 IEEE International
  Conference on Robotics and Automation (ICRA)}.\hskip 1em plus 0.5em minus
  0.4em\relax IEEE, 2018, pp. 86--92.

\bibitem{ito2018small}
S.~Ito, S.~Hiratsuka, M.~Ohta, H.~Matsubara, and M.~Ogawa, ``Small imaging
  depth lidar and dcnn-based localization for automated guided vehicle,''
  \emph{Sensors}, vol.~18, no.~1, p. 177, 2018.

\bibitem{hess2016real}
W.~Hess, D.~Kohler, H.~Rapp, and D.~Andor, ``Real-time loop closure in 2d lidar
  slam,'' in \emph{2016 IEEE International Conference on Robotics and
  Automation (ICRA)}.\hskip 1em plus 0.5em minus 0.4em\relax IEEE, 2016, pp.
  1271--1278.

\bibitem{li2014lidar}
R.~Li, J.~Liu, L.~Zhang, and Y.~Hang, ``Lidar/mems imu integrated navigation
  (slam) method for a small uav in indoor environments,'' in \emph{2014 DGON
  Inertial Sensors and Systems (ISS)}.\hskip 1em plus 0.5em minus 0.4em\relax
  IEEE, 2014, pp. 1--15.

\bibitem{besl1992method}
P.~J. Besl and N.~D. McKay, ``Method for registration of 3-d shapes,'' in
  \emph{Sensor fusion IV: control paradigms and data structures}, vol.
  1611.\hskip 1em plus 0.5em minus 0.4em\relax International Society for Optics
  and Photonics, 1992, pp. 586--606.

\bibitem{zhang2017low}
J.~Zhang and S.~Singh, ``Low-drift and real-time lidar odometry and mapping,''
  \emph{Autonomous Robots}, vol.~41, no.~2, pp. 401--416, 2017.

\bibitem{deschaud2018imls}
J.-E. Deschaud, ``Imls-slam: scan-to-model matching based on 3d data,'' in
  \emph{2018 IEEE International Conference on Robotics and Automation
  (ICRA)}.\hskip 1em plus 0.5em minus 0.4em\relax IEEE, 2018, pp. 2480--2485.

\bibitem{zhang2014loam}
J.~Zhang and S.~Singh, ``Loam: Lidar odometry and mapping in real-time.'' in
  \emph{Robotics: Science and Systems}, vol.~2, no.~9, 2014.

\bibitem{shan2018lego}
T.~Shan and B.~Englot, ``Lego-loam: Lightweight and ground-optimized lidar
  odometry and mapping on variable terrain,'' in \emph{2018 IEEE/RSJ
  International Conference on Intelligent Robots and Systems (IROS)}, 2018, pp.
  4758--4765.

\bibitem{rozenberszki2020lol}
D.~Rozenberszki and A.~L. Majdik, ``Lol: Lidar-only odometry and localization
  in 3d point cloud maps,'' in \emph{2020 IEEE International Conference on
  Robotics and Automation (ICRA)}.\hskip 1em plus 0.5em minus 0.4em\relax IEEE,
  2020, pp. 4379--4385.

\bibitem{liosam2020shan}
T.~Shan, B.~Englot, D.~Meyers, W.~Wang, C.~Ratti, and R.~Daniela, ``Lio-sam:
  Tightly-coupled lidar inertial odometry via smoothing and mapping,'' in
  \emph{IEEE/RSJ International Conference on Intelligent Robots and Systems
  (IROS)}.\hskip 1em plus 0.5em minus 0.4em\relax IEEE, 2020.

\bibitem{yin2020cae}
D.~Yin, Q.~Zhang, J.~Liu, X.~Liang, Y.~Wang, J.~Maanp{\"a}{\"a}, H.~Ma,
  J.~Hyypp{\"a}, and R.~Chen, ``Cae-lo: Lidar odometry leveraging fully
  unsupervised convolutional auto-encoder for interest point detection and
  feature description,'' \emph{arXiv preprint arXiv:2001.01354}, 2020.

\bibitem{masci2011stacked}
J.~Masci, U.~Meier, D.~Cire{\c{s}}an, and J.~Schmidhuber, ``Stacked
  convolutional auto-encoders for hierarchical feature extraction,'' in
  \emph{International conference on artificial neural networks}.\hskip 1em plus
  0.5em minus 0.4em\relax Springer, 2011, pp. 52--59.

\bibitem{shi2020spsequencenet}
H.~Shi, G.~Lin, H.~Wang, T.-Y. Hung, and Z.~Wang, ``Spsequencenet: Semantic
  segmentation network on 4d point clouds,'' in \emph{Proceedings of the
  IEEE/CVF Conference on Computer Vision and Pattern Recognition}, 2020, pp.
  4574--4583.

\bibitem{bergelt2017improving}
R.~Bergelt, O.~Khan, and W.~Hardt, ``Improving the intrinsic calibration of a
  velodyne lidar sensor,'' in \emph{2017 IEEE SENSORS}.\hskip 1em plus 0.5em
  minus 0.4em\relax IEEE, 2017, pp. 1--3.

\bibitem{kirillov2008introduction}
A.~Kirillov~Jr, \emph{An introduction to Lie groups and Lie algebras}.\hskip
  1em plus 0.5em minus 0.4em\relax Cambridge University Press, 2008, vol. 113.

\bibitem{zhang2017robust}
L.~Zhang and P.~N. Suganthan, ``Robust visual tracking via co-trained
  kernelized correlation filters,'' \emph{Pattern Recognition}, vol.~69, pp.
  82--93, 2017.

\bibitem{barfoot2016state}
T.~D. Barfoot, ``State estimation for robotics: A matrix lie group approach,''
  \emph{Draft in preparation for publication by Cambridge University Press,
  Cambridge}, 2016.

\bibitem{rusu20113d}
R.~B. Rusu and S.~Cousins, ``3d is here: Point cloud library (pcl),'' in
  \emph{2011 IEEE international conference on robotics and automation}.\hskip
  1em plus 0.5em minus 0.4em\relax IEEE, 2011, pp. 1--4.

\bibitem{geiger2013vision}
A.~Geiger, P.~Lenz, C.~Stiller, and R.~Urtasun, ``Vision meets robotics: The
  kitti dataset,'' \emph{The International Journal of Robotics Research},
  vol.~32, no.~11, pp. 1231--1237, 2013.

\bibitem{Geiger2012CVPR}
A.~Geiger, P.~Lenz, and R.~Urtasun, ``Are we ready for autonomous driving? the
  kitti vision benchmark suite,'' in \emph{Conference on Computer Vision and
  Pattern Recognition (CVPR)}, 2012.

\bibitem{mur2015orb}
R.~Mur-Artal, J.~M.~M. Montiel, and J.~D. Tardos, ``Orb-slam: a versatile and
  accurate monocular slam system,'' \emph{IEEE transactions on robotics},
  vol.~31, no.~5, pp. 1147--1163, 2015.

\bibitem{qin2019general}
T.~Qin, J.~Pan, S.~Cao, and S.~Shen, ``A general optimization-based framework
  for local odometry estimation with multiple sensors,'' \emph{arXiv preprint
  arXiv:1901.03638}, 2019.

\bibitem{graeter2018limo}
J.~Graeter, A.~Wilczynski, and M.~Lauer, ``Limo: Lidar-monocular visual
  odometry,'' in \emph{2018 IEEE/RSJ International Conference on Intelligent
  Robots and Systems (IROS)}.\hskip 1em plus 0.5em minus 0.4em\relax IEEE,
  2018, pp. 7872--7879.

\bibitem{engel2015large}
J.~Engel, J.~St{\"u}ckler, and D.~Cremers, ``Large-scale direct slam with
  stereo cameras,'' in \emph{Intelligent Robots and Systems (IROS), 2015
  IEEE/RSJ International Conference on}.\hskip 1em plus 0.5em minus 0.4em\relax
  IEEE, 2015, pp. 1935--1942.

\bibitem{koide2019portable}
K.~Koide, J.~Miura, and E.~Menegatti, ``A portable three-dimensional
  lidar-based system for long-term and wide-area people behavior measurement,''
  \emph{International Journal of Advanced Robotic Systems}, vol.~16, no.~2, p.
  1729881419841532, 2019.

\bibitem{qin2018vins}
T.~Qin, P.~Li, and S.~Shen, ``Vins-mono: A robust and versatile monocular
  visual-inertial state estimator,'' \emph{IEEE Transactions on Robotics},
  vol.~34, no.~4, pp. 1004--1020, 2018.

\bibitem{o2014gentle}
J.~M. O'Kane, ``A gentle introduction to ros,'' 2014.

\end{thebibliography}

\end{document}